\documentclass[conference]{IEEEtran}
\usepackage[paper=letterpaper, top=0.75in, bottom=1in, left=0.67in, right=0.67in]{geometry}
\IEEEoverridecommandlockouts

\usepackage{cite}
\usepackage{nicefrac}
\usepackage{url}
\usepackage{amsmath,amssymb,amsfonts, amsthm}
\usepackage{algorithm}
\usepackage{algorithmic}
\usepackage{subcaption}
\usepackage{graphicx}
\usepackage{textcomp}
\usepackage{xcolor}
\usepackage[draft]{hyperref}

\usepackage{cleveref} 
\usepackage{epstopdf}
\def\BibTeX{{\rm B\kern-.05em{\sc i\kern-.025em b}\kern-.08em
    T\kern-.1667em\lower.7ex\hbox{E}\kern-.125emX}}
\begin{document}

\title{Optimisation of Resource Allocation in Heterogeneous Wireless Networks Using Deep Reinforcement Learning
\thanks{This work was supported by the African Institute for Mathematical Sciences (AIMS), South Africa, and the Mastercard Foundation Scholarship.\newline
The work of Tobi Awodumila has received funding from the Google DeepMind Scholarship under AIMS.\newline
}
}

\author{\IEEEauthorblockN{Oluwaseyi Giwa\IEEEauthorrefmark{1}, Jonathan Shock\IEEEauthorrefmark{2}, Jaco Du Toit\IEEEauthorrefmark{3}, and Tobi Awodumila\IEEEauthorrefmark{1}}
\IEEEauthorblockA{\IEEEauthorrefmark{1}African Institute for Mathematical Sciences, South Africa}
\IEEEauthorblockA{\IEEEauthorrefmark{2}University of Cape Town, South Africa}
\IEEEauthorblockA{\IEEEauthorrefmark{3}Vodacom, South Africa}
\IEEEauthorblockA{Email: \{oluwaseyi, tobi\}@aims.ac.za, jonathan.shock@uct.ac.za, jacowp357@gmail.com}
}

\maketitle

\begin{abstract}
Dynamic resource allocation in open radio access network (O-RAN) heterogeneous networks (HetNets) presents a complex optimisation challenge under varying user loads. We propose a near-real-time RAN intelligent controller (Near-RT RIC) xApp utilising deep reinforcement learning (DRL) to jointly optimise transmit power, bandwidth slicing, and user scheduling. Leveraging real-world network topologies, we benchmark proximal policy optimisation (PPO) and twin delayed deep deterministic policy gradient (TD3) against standard heuristics. Our results demonstrate that the PPO-based xApp achieves a superior trade-off, reducing network energy consumption by up to \(70\%\) in dense scenarios and improving user fairness by more than \(30\%\) compared to throughput-greedy baselines. These findings validate the feasibility of centralised, energy-aware AI orchestration in future 6G architectures.
\end{abstract}


\section{Introduction}\label{intro}
The evolution towards fifth-generation (5G) and the forthcoming sixth-generation (6G) wireless systems is driven by a demand for ubiquitous connectivity and high data rates. This has led to the proliferation of heterogeneous networks (HetNets), which overlay traditional macrocells with dense tiers of small cells (e.g., micro, pico, and femto cells) to enhance spectral efficiency and network capacity~\cite{xu2021, faeq2023}. However, this architectural complexity introduces challenges in resource allocation (RA). The dense deployment of base stations (BS) creates severe co-tier and cross-tier interference, making the efficient management of spectrum, transmit power, and user association critical for network performance. Optimising these resources is essential not only to maximise throughput but also to ensure fairness and quality of service (QoS) for all users in the network~\cite{agarwal2022, ather2025}.

\subsection{Related Works}
Traditional RA strategies, relying on classical optimisation or heuristics~\cite{boyd2004}, are inadequate for modern HetNets~\cite{mughees2023}. These methods depend on simplified, static network models and struggle with the nonconvex, combinatorial nature of joint RA problems. Furthermore, distributed approaches often lack the global view necessary for optimal interference coordination. The emergence of open radio access networks (O-RAN) addresses this by introducing the near-real-time RAN intelligent controller (Near-RT RIC), which enables centralised, data-driven control via xApps~\cite{oluwaseyi2025the}.

Reinforcement learning (RL) has emerged as a powerful paradigm for this challenge. By learning policies through direct interaction with the environment~\cite {sutton1998}, RL agents adapt to real-time conditions without an explicit model. Recent deep reinforcement learning (DRL) approaches effectively handle the high-dimensional state and action spaces of modern networks~\cite{mnih2013, hasselt2016, haarnoja2019, archi2023, tian2023, chi2024, park2024, olayemi2024, shalini2024}, demonstrating superior performance over rule-based methods in tasks ranging from power control to network slicing.

Recent literature has increasingly explored DRL in wireless networks. For instance, variants of deep-Q-networks (DQN)~\cite{chi2024}, and deep deterministic policy gradient (DDPG)/twin delayed deep deterministic policy gradient (TD3)~\cite{olayemi2024} have shown promise in computation offloading and autonomous navigation, while decentralized multi-agent learning is gaining traction for dynamic resource management~\cite{shalini2024}. However, applying these advanced DRL frameworks specifically within the constraints of an O-RAN architecture remains an open challenge.

\subsection{Contributions}
While RL for RA is well-investigated, existing work often relies on simplified synthetic topologies or isolates power control from scheduling. This paper bridges the gap between theoretical DRL and realistic deployment constraints. Our specific contributions include formulating a Near-RT RIC-compatible Markov decision process (MDP) in which a central agent manages power and scheduling using global channel knowledge, justified via O-RAN E2 feedback loops. Second, we implement a simulation environment using real-world BS coordinates to capture realistic interference geometries, unlike purely Poisson Point Process (PPP) models. Finally, we provide a mathematical derivation of throughput and fairness metrics from continuous RL actions, comparing TD3 and proximal policy optimisation (PPO) against standard heuristics. Simulation results show that DRL agents outperform heuristic baselines by over \(\sim 70\%\) in energy reduction and \(\geq 100\%\) in throughput while maintaining better fairness among users. The remainder of this paper is organised as follows: Section~\ref{model} details the system model and problem formulation. Section~\ref{algorithms} describes the DRL algorithms. Section~\ref{experiment} presents the experimental setup. Section~\ref{results} discusses the results, and Section~\ref{conclusion} concludes the paper.

\section{System Model}\label{model}
We consider a downlink HetNet operating within an O-RAN architecture. The network consists of a set of BSs, \(B = \{1, \dots, N_B\}\), comprising \(N_M\) macro BSs and \(N_S\) micro BSs. These serve a set of user equipments (UEs) \(U = \{1, \dots, N_U\}\) distributed stochastically within the coverage area.

\begin{figure}[t]
    \centering
    \includegraphics[width=0.8\columnwidth]{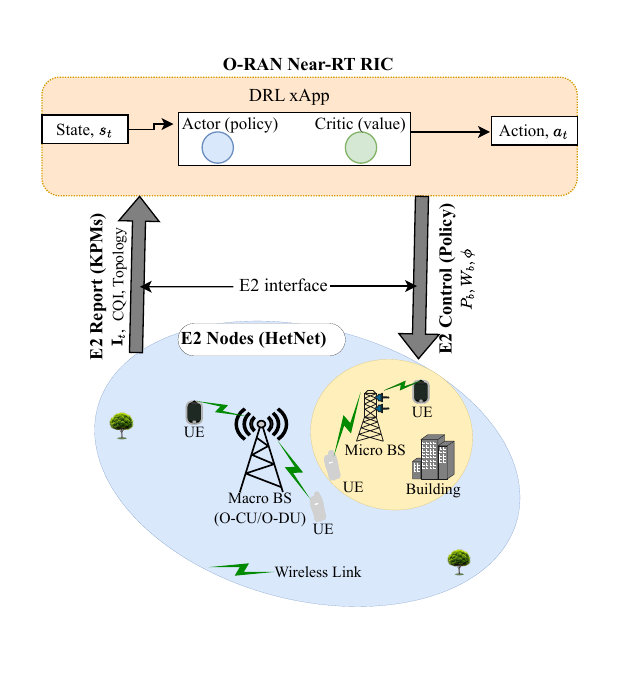}
    \caption{The proposed O-RAN compliant architecture. The Deep RL agent operates as an xApp within the Near-RT RIC, collecting KPMs via the E2 interface to construct the state \(s_t\) and issuing optimizing control policies \(a_t\) to the Macro and Micro E2 nodes.}
    \label{fig:rl_diagram}
\end{figure}

The system is controlled by a centralised Near-RT RIC that hosts an xApp responsible for optimising radio resources at discrete time intervals \(t\) (cf Fig.~\ref{fig:rl_diagram}).

\subsection{Channel Model and Signal Quality}\label{subsec:channel-model}
Let \(p_{b, t}\) denote the transmit power of BS \(b\) at time \(t\), and \(x_{b, u} \in \{0, 1\}\) be the binary association variable, where \(x_{b, u} = 1\) if user \(u\) is served by BS \(b\).\\
The wireless channel between BS \(b\) and user \(u\) accounts for path loss, log-normal shadowing, and fast fading. The received power \(P_{u, b}^{rx}\) is given by:
\begin{equation}
    P_{u, b}^{rx} = p_{b, t} \cdot H_{b, u} \cdot \zeta_{b, u}(t),
\end{equation}
where \(H_{b, u} = d_{b, u}^{-\eta}10^{\frac{\xi_{b, u}}{10}}\) represents the large-scale channel gain (distance-dependent path loss with exponent \(\eta\) and shadowing \(\xi_{b, u} \sim \mathcal{N}(0, \sigma_{sh}^{2})\)). The term \(\zeta_{b, u}(t)\) represents the small-scale Rayleigh fading component, assumed to be unit-mean exponential random variables.

The Signal-to-Interference-plus-Noise Ratio (SINR) for user \(u\) associated with BS \(b\) is formulated as:
\begin{equation}
    \rm{SINR}_{u, b}(t) = \frac{p_{b, t} H_{b, u} \zeta_{b, u}(t)}{\sum_{j \in \mathcal{B} \setminus {b}}p_{j, t}H_{j, u} \zeta_{j, u}(t) + N_{0} W},
    \label{eq:sinr-formulation}
\end{equation}
where \(N_{0}\) is the noise spectral density and \(W\) is the system bandwidth.

\subsection{Throughput and Energy Metrics}
The available bandwidth at BS \(b\), denoted as \(W_{b} \in [0, W]\), is dynamically adjusted to mitigate interference. The scheduler at BS \(b\) allocates a fraction \(\phi_{u, b}(t)\) of \(W_b\) to user \(u\), such that \(\sum_{u \in U_{b}} \phi_{u, b}(t) \leq 1\). The achievable data rate for user \(u\) is given by the Shannon capacity:
\begin{equation}
    R_{u}(t) = \sum_{b \in \mathcal{B}} x_{b, u} \cdot \phi_{u, b}(t) W_{b}(t) \log_{2}\left(1 + \rm{SINR}_{u, b}(t)\right).
    \label{eq:data-rate}
\end{equation}
We strictly define the network energy consumption \(E_{\rm{net}}(t)\) as the sum of radiated power:
\begin{equation}
    E_{\rm{net}}(t) = \sum_{b \in B} p_{b, t}.
    \label{eq:energy-consumption}
\end{equation}
To quantify user fairness, we utilise Jain's Fairness Index \(\mathcal{J}(t)\), defined over the rate vector \(R(t) = [R_{1}(t), \dots, R_{N_U}(t)]\):
\begin{equation}
    \mathcal{J}(\mathbf{R}(t)) = \frac{\left(\sum_{u = 1}^{N_U} R_{u(t)}\right)^2}{N_U \sum_{u = 1}^{N_U} R_{u}(t)^2}.
    \label{eq:fairness-index}
\end{equation}

\subsection{Optimisation Problem}
The objective is to find a joint policy \(\pi\) for power control \(\mathbf{p}\), bandwidth slicing \(\mathbf{W}\), and scheduling weights \(\boldsymbol{\phi}\) that maximises a multi-objective utility function over a horizon \(T\). This creates a non-convex, combinatorial optimisation problem:
\begin{align}
    \max_{\mathbf{p}, \mathbf{W}, \boldsymbol{\phi}} \quad & \lim_{T \to \infty} \frac{1}{T}\sum_{t = 0}^{T} \left[\omega_1 \sum_u R_u(t) + \omega_2 \mathcal{J}(t) - \omega_3 E_{\rm{net}}(t)\right] \tag{P1}\\
    \text{s.t.} \quad & 0 \leq p_{b, t} \leq P_{\rm{max}}, \quad \forall b \in B \notag \\
    & 0 \leq W_{b}(t) \leq W, \quad \forall b \in B \notag
\end{align}
Direct solution of (P1) is intractable due to the coupling of interference in SINR \eqref{eq:sinr-formulation} and the continuous-discrete nature of variables.

\subsection{MDP Formulation for O-RAN xApp}
To solve (P1), we formulate the problem as a MDP \((\mathcal{S}, \mathcal{A}, \mathcal{P}, \mathcal{R})\). The agent (xApp) interacts with the environment (E2 nodes) as follows:\\
\textbf{State Space \(\mathcal{S}\) (xApp input)}: The state \(s_t\) aggregates global network observables available at the RIC via E2 key performance measurement (KPM) reports:
\begin{equation}
    s_t = \left\{\mathbf{p}_{t-1}, \{\mathbf{I}_{u}^{\rm{est}}\}_{u \in U}, \mathbf{L}_{\rm{geo}}\right\},
\end{equation}
where \(\mathbf{p}_{t - 1}\) is the previous power allocation, \(\mathbf{I}_{u}^{\rm{est}}\) is the estimated interference measurement from UE channel quality indicator (CQI) reports, and \(\mathbf{L}_{\rm{geo}}\) encapsulates the fixed topology geometry.\\
\textbf{Hierarchical Action Space \(\mathcal{A}\) (xApp output)}: To bridge the timescale gap between RIC (approx. 10ms - 1s) and medium access control (MAC) scheduling (1ms), the agent learns high-level policy parameters rather than instantaneous scheduling decisions. The action vector \(a_t \in [-1, 1]^{3N_{B}}\) consists of normalised values mapped to physical quantities:
\begin{itemize}
    \item Power Control \((\hat{p}_b)\): Scaled to \(p_{b, t} \in [P_{\rm{min}}, P_{\rm{max}}]\).
    \item Bandwidth Slice \((\hat{w}_b)\): Scaled to \(W_{b}(t) \in [0, W]\).
    \item User Priority Weight \((\hat{psi}_{b, u})\): This scalar influences the local scheduler. The actual resource fraction \(\phi_{u, b}\) is derived via a softmax function to ensure validity and differentiability:
    \begin{equation}
        \phi_{u, b}(t) = \frac{e^{(\tau \cdot \hat{\psi}_{b, u})}}{\sum_{k \in U_b}e^{(\tau \cdot \hat{\psi}_{b, k})}},
    \end{equation}
\end{itemize}
where \(\tau\) is a temperature parameter. This effectively enables the RL agent to bias the local proportional fair scheduler towards specific users (e.g., cell-edge) to enforce fairness.\\
\textbf{Reward Function \(\mathcal{R}\)}: The reward \(r_t\) is a direct scalarisation of the objective in (P1):
\begin{equation}
    r_t = \alpha \frac{\sum R_u(t)}{R_{\rm{max}}} + \beta \mathcal{J}(t) - \kappa \frac{\sum p_{b, t}}{N_B P_{\rm{max}}},
    \label{eq:reward-function}
\end{equation}
where coefficients \(\alpha, \beta, \kappa\) prioritise throughput, fairness, and energy efficiency, respectively. Normalisation terms ensure numerical stability during gradient descent.

\begin{figure}[t]
    \centering
    \includegraphics[width=1.0\columnwidth]{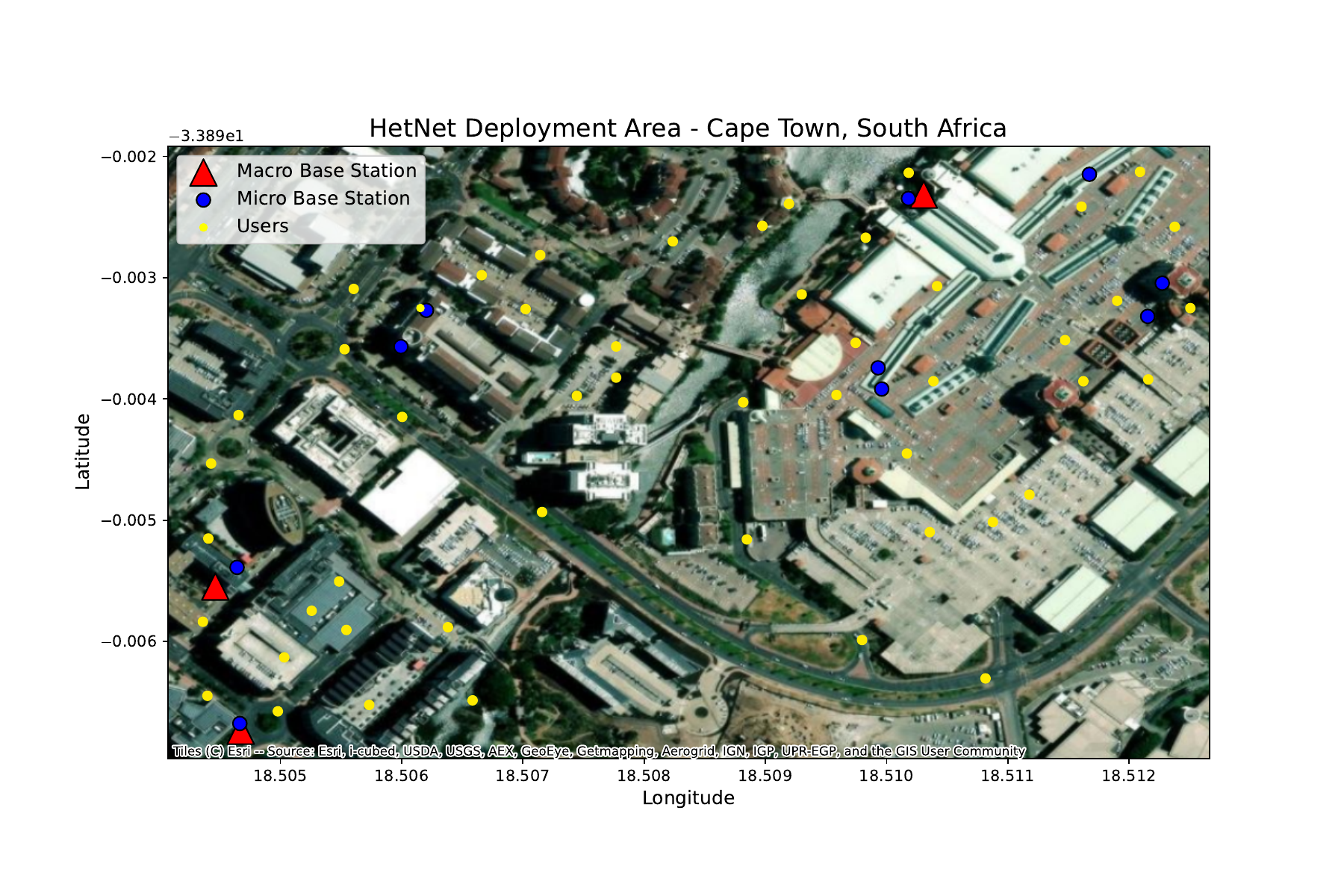}
    \caption{Satellite image of the deployment area. Source: Esri GIS software for mapping and spatial analysis.}
    \label{fig:hetnet_map}
\end{figure}

\section{DRL Algorithms}\label{algorithms}
The RA problem formulated in Section~\ref{model} is characterised by a high-dimensional state space and a continuous action space (for transmit power and bandwidth). This renders DRL algorithms, such as DQN, which are restricted to discrete actions, unsuitable. Consequently, we turn to actor-critic and policy-gradient methods, which are designed for continuous control. While DDPG is a natural starting point, it is known to suffer from instability and overestimation of Q-values. We therefore select two state-of-the-art algorithms that address these challenges: TD3, which directly mitigates DDPG's shortcomings, and PPO, renowned for its robustness and stable training performance.

\subsection{Twin Delayed Deep Deterministic Policy Gradient (TD3)}
TD3 is an off-policy, model-free algorithm that builds upon DDPG by introducing several key modifications to enhance stability and performance (cf Alg.~\ref{alg:td3}). It learns a deterministic policy (the actor) that maps states to actions, and a Q-function (the critic) that estimates the action-value function. The three core innovations of TD3 are:

\textbf{Clipped Double Q-Learning:} To combat the overestimation bias of the critic, TD3 employs two independent critic networks, \(Q_{\theta_1}\) and \(Q_{\theta_2}\). When computing the target value for the Bellman update, it takes the minimum of the two critics' predictions, yielding a more conservative and stable target:
\begin{equation}
    y = r + \gamma \min_{i = 1, 2} Q_{\theta_{i}'}\left(s', \pi_{\mu'}(s') + \epsilon\right)
\end{equation}
Where \(\mu' \; \text{and} \; \theta'\) are the parameters of the target networks, and the noise \(\epsilon\) is for target policy smoothing.

\textbf{Delayed Policy Updates:} The actor network (\(\pi_{\mu}\)) is updated less frequently than the critic networks. This allows the critic's Q-value estimates to converge and stabilise before being used to update the actor, leading to more reliable policy improvements.

\textbf{Target Policy Smoothing:} Noise is added to the target action during the target Q-value calculation. This helps regularise the policy, making it less likely to exploit narrow peaks in the value function, resulting in a smoother policy landscape.

\begin{algorithm}[t]
    \caption{TD3 for Resource Allocation Optimisation}
    \label{alg:td3}
    \begin{algorithmic}[1]
        \STATE \textbf{Initialise} actor \(\pi_{\mu}\), critics \(Q_{\theta_1}\), \(Q_{\theta_2}\), and their target networks \(\pi_{\mu'}, Q_{\theta_{1}'}, Q_{\theta_{2}'}\) and replay buffer \(\mathcal{D}\).
        \FOR{each training step}
            \STATE Select action with exploration noise:
            \(a = \pi_{\mu}(s) + \mathcal{N}(0, \sigma)\).
            \STATE Store \((s, a, r, s')\) in \(\mathcal{D}\) and sample a minibatch from \(\mathcal{D}\)
            \STATE Compute target action with smoothed noise: \(
            a' \gets \pi_{\mu'}(s') + \text{clip}\left(\mathcal{N(0, \sigma}), -c, c\right)
            \).
            \STATE Compute target Q-value:
            \(
            y = r + \gamma \min_{i = 1, 2} Q_{\theta_{i}'}\left(s', a'\right)
            \)
            \STATE Update critics \(\theta_i\) by minimising Huber/MSE loss:
            \(\mathcal{L}(\theta_i) = \left(Q_{\theta_i}(s, a) - y\right)^2\).
            \IF{step is a policy update step}
                \STATE Softly update all target networks: \(
                \theta' \gets \tau\theta + (1 - \tau)\theta',  \mu' \gets \tau\mu + (1 - \tau)\mu'\).
            \ENDIF
        \ENDFOR
    \end{algorithmic}
\end{algorithm}

\subsection{Proximal Policy Optimisation (PPO)}
PPO is an on-policy actor-critic algorithm known for its balance between sample efficiency and ease of implementation. Unlike TD3, PPO learns a stochastic policy, \(\pi_{\theta}(a|s)\). Its key feature is a novel surrogate objective function that constrains the size of policy updates, preventing destructive, large changes during training.
The core of PPO is the clipped surrogate objective, which modifies the standard policy-gradient objective (cf Alg.~\ref{alg:ppo}). It uses the ratio between the new policy and the old policy, \(r_t(\theta) = \frac{\pi_{\theta}(a_t|s_t)}{\pi_{\theta_{\text{old}}}(a_t|s_t)}\), to measure the policy change. The objective is:
\begin{equation}
    L^{\text{CLIP}}(\theta) = \hat{\mathbb{E}}_{t}\left[\min\left(r_t(\theta)\hat{A_t}, \text{clip}\left(r_t(\theta), 1 - \epsilon, 1 + \epsilon\right)\hat{A_t}\right)\right]
\end{equation}
Where \(\hat{A_t}\) is an estimate of the advantage function (often computed using generalised advantage estimation, GAE), and \(\epsilon\) is a small hyperparameter that defines the clipping range. This objective clips the probability ratio, which discourages policy updates that move \(r_t(\theta)\) far from 1, thereby ensuring more stable training.

\begin{algorithm}[t]
    \caption{PPO for Resource Allocation Optimisation}
    \label{alg:ppo}
    \begin{algorithmic}[1]
        \STATE \textbf{Initialise} actor-critic network parameters \(\theta\).
        \FOR{each iteration}
            \STATE Collect a set of trajectories by running policy \(\pi_{\theta_{\text{old}}}\) in the environment for \(T\) timesteps.
            \STATE Compute advantage estimates \(\hat{A_1}, \dots, \hat{A_T}\) (using GAE).
            \FOR{a fixed number of epochs}
                \STATE Optimise the surrogate objective on the collected data via stochastic gradient ascent: \(\theta \gets \theta + \alpha\nabla_{\theta}L^{\text{CLIP}}(\theta)\)
            \ENDFOR
            \STATE \(\theta_{\text{old}} \gets \theta\).
        \ENDFOR
    \end{algorithmic}
\end{algorithm}

\section{Experimental Scenarios and Setup}\label{experiment}
\subsection{Simulation Environment and Topology}
We developed a custom O-RAN-compliant simulation environment to evaluate the proposed RIC xApp.\\
\textbf{Topology}: The network layout is instantiated using real-world BS geospatial data from a telecom operator in Cape Town, South Africa. The dataset comprises \(N_M = 3\) macro BSs and \(N_S = 10\) micro BSs. While BS locations are fixed to preserve realistic interference geometries, \(N_U = 50\) users are randomly distributed within the deployment polygon at the start of each episode to ensure the policy generalises across spatial distributions. Fig.~\ref{fig:hetnet_map} shows the satellite view used to derive the layout. Colors in all figures follow the evaluation convention: \textcolor{red}{Macro BS (red)}, \textcolor{blue}{Micro BS (blue)}, \textcolor{yellow!80!black}{Users (yellow)}.\\
\textbf{Channel Parameters}: The channel propagation follows the model defined in Section~\ref{subsec:channel-model}. While the network layout leverages real-world geospatial coordinates to preserve realistic interference geometries, we utilise standardised constant path-loss (\(\eta = 3.5\)) and shadowing (\(\sigma_{sh} = 8\) dB) parameters. This ensures our DRL agents can be benchmarked objectively against widely accepted theoretical channel conditions, rather than overfitting to a specific operator's proprietary RF measurement data. The small-scale fading \(\zeta_{b, u}(t)\) is modelled as independent and identically distributed (i.i.d.) Rayleigh fading, with a new random variable drawn at each transmission time interval (TTI) to accurately capture instantaneous fast channel variations. The system bandwidth is \(W = 20\) MHz, and thermal noise density is \(N_0 = -174\) dBm/Hz.

\subsection{Action Mapping and Hyperparameters}
The RL agent's normalised actions \(a_t \in [-1, 1]\) are mapped to physical resources as follows:\\
\textbf{Power}: Transmit power \(p_{b,t}\) is scaled linearly. We set \(P_{\rm{max}}\) to 46 dBm for macro BSs and 30 dBm for micro BSs, with a dynamic range of 20 dB.\\
\textbf{Scheduling}: The softmax temperature parameter is set to \(\tau = 1.0\), allowing the agent to smoothly transition between strict priority scheduling and round-robin behaviour.

\subsection{Training and Evaluation}
We train TD3 and PPO agents over 1000 episodes with a horizon of \(T = 1000\) steps per episode. The reward function weights in \eqref{eq:reward-function} are tuned via grid searches to \(\alpha = 1.0, \; \beta = 2.0, \; \text{and, } \kappa = 0.5\), prioritising equitable service coverage. We compare the DRL agents against three baselines: \textbf{(1) Greedy OFDMA (G-OFDMA)}: assigns RB to the user with the best SINR, \textbf{(2) Interference Pricing (IP-PC)}: reduces power based on neighbour feedback, and \textbf{(3) Proportional Fair (PF-EQ)}: standard baseline for fairness.

\subsection{Performance Metrics}
To evaluate the proposed O-RAN xApp against the baselines, we assess the trained policies on a hold-out test set using the following physical key performance indicators (KPIs):

\textbf{Average Per-User Throughput \((\bar{R}_{avg})\)}: This metric quantifies the mean data rate available to an individual user, serving as a primary indicator of Quality of Service (QoS). It is calculated by averaging the instantaneous rates \eqref{eq:data-rate} across all users and time steps:
\begin{equation}
    \bar{R}_{avg} = \frac{1}{T_{eval} \cdot N_U} \sum_{t = 1}^{T_{eval}} \sum_{u = 1}^{N_U} R_u(t) \quad [\text{Mbps}].
\end{equation}

\textbf{Average Fairness Index \((\bar{\mathcal{J}})\)}: To ensure the policy does not maximise throughput by starving cell-edge users, we report the time-averaged Jain's fairness index. This corresponds to the stability of the fairness objective defined in \eqref{eq:fairness-index}:
\begin{equation}
    \bar{\mathcal{J}} = \frac{1}{T_{eval}} \sum_{t = 1}^{T_{eval}} \mathcal{J}(\mathbf{R}(t)).
\end{equation}
A value closer to 1 indicates an equitable distribution of resources among all users, regardless of their channel conditions.

\textbf{Network Energy Consumption \((\bar{E}_{\rm{net}})\)}: We evaluate the environmental impact of the xApp by measuring the average aggregate transmission power of the network, derived from \eqref{eq:energy-consumption}:
\begin{equation}
    \bar{E}_{\rm{net}} = \frac{1}{T_{eval}}\sum_{t = 1}^{T_{eval}} \sum_{b = 1}^{N_B} p_{b, t} \quad [\text{Watts}].
\end{equation}
Lower values indicate that the agent successfully learns to mitigate interference by reducing power rather than simply increasing it.

\textbf{Average Reward \((\bar{r})\)}: For the DRL agents, we track the cumulative reward per episode to analyse convergence behaviour and sample efficiency. This serves as a holistic metric of how well the agent balances the conflicting objectives of throughput, fairness, and energy, as defined in \eqref{eq:reward-function}.

\section{Performance Comparison and Discussion}\label{results}
We evaluate the proposed O-RAN xApps (PPO and TD3) against heuristics across four topological scenarios: Dense Urban (\(10 N_S, 3 N_M\), high interference), Sparse Suburban (\(3 N_M\) only), Hotspot (users cluster near \(N_S\)), and Mixed (random \(N_S\) plus uniform users). The analysis focuses on the trade-offs between the conflicting objectives defined in Section~\ref{model}

\subsection{Throughput - Energy Trade-off}
The trade-off between spectral efficiency and green networking is evident in the Dense Urban and Hotspot scenarios (Figs.~\ref{fig:dense-urban} and~\ref{fig:hotspot-scenario}). G-OFDMA achieves a competitive average throughput, but at the cost of maximum energy consumption (normalised \(\bar{E}_{\rm{net}} \approx 0.95 - 1.0\)). Ignoring inter-cell interference forces all BSs to transmit at peak power. IP-PC successfully minimises energy (\(\bar{E}_{\rm{net}} \approx 0.15\)) but results in the lowest user throughput due to overly aggressive power back-off in response to interference pricing.\\

PPO xApp strikes an optimal balance. In the Dense Urban scenario, PPO achieves a \(\sim 70\%\) reduction in energy consumption compared to G-OFDMA while maintaining superior per-user throughput (\(\bar{R}_{avg}\)). This confirms the agent effectively learns to utilise the ``silent'' periods and power control actions (\(\hat{p}_b\)) to mitigate cross-tier interference, maximising the SINR rather than just the signal power.

\subsection{Fairness and QoS Assurance}
This is a critical requirement for 6G O-RAN to ensure equitable Service Level Agreements (SLAs). Across all scenarios, G-OFDMA yields poor fairness (\(\bar{\mathcal{J}} < 0.25\)), indicating that cell-edge users are starved to maximise the sum-rate of cell-centre users. PPO demonstrates superior fairness management, achieving a Jain’s Index of \(\approx 0.65 - 0.75\) in all topologies. Notably, in the Mixed Scenario (Fig.\ref{fig:mixed-scenario}), PPO improves fairness by over \(300\%\) compared to G-OFDMA and \(100\%\) compared to IP-PC. While PF-EQ is designed for fairness, it lacks the interference coordination capabilities of the global xApp, resulting in significantly lower aggregate throughput than the DRL agents.

\subsection{Learning Convergence and Computational Complexity}
\begin{figure*}[t]
    \centering
    \begin{subfigure}{0.32\textwidth}
        \includegraphics[width=\textwidth]{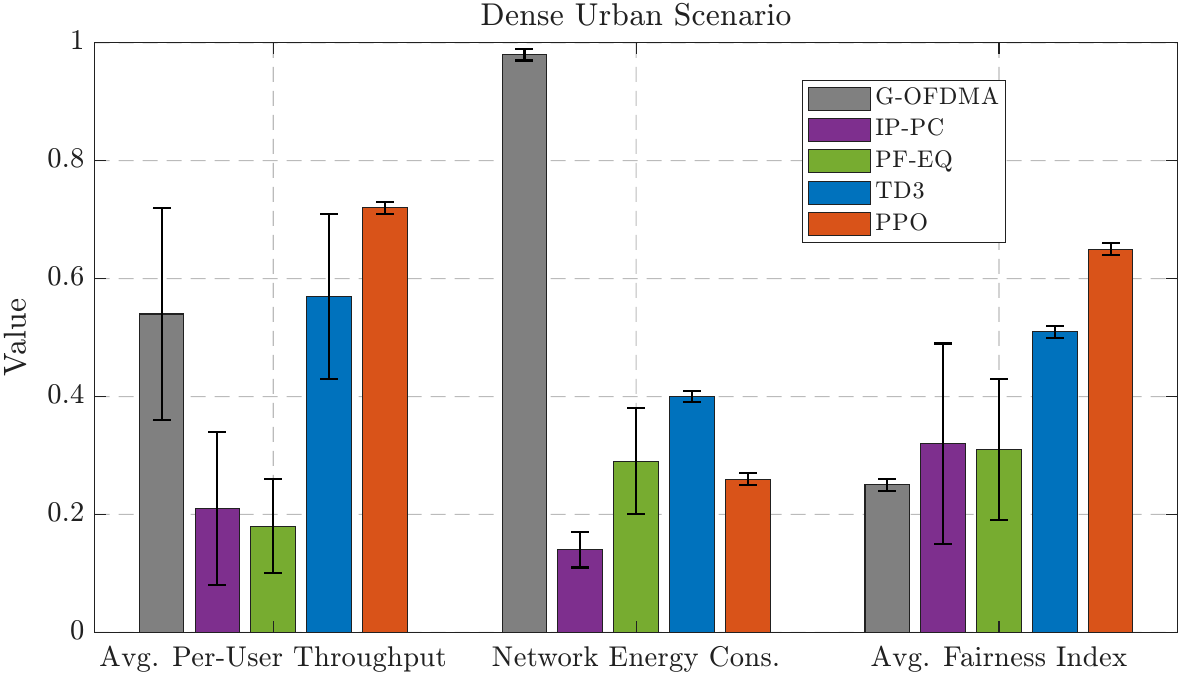}
        \caption{}
        \label{fig:dense-urban}
    \end{subfigure}
    \begin{subfigure}{0.32\textwidth}
        \centering
        \includegraphics[width=\textwidth]{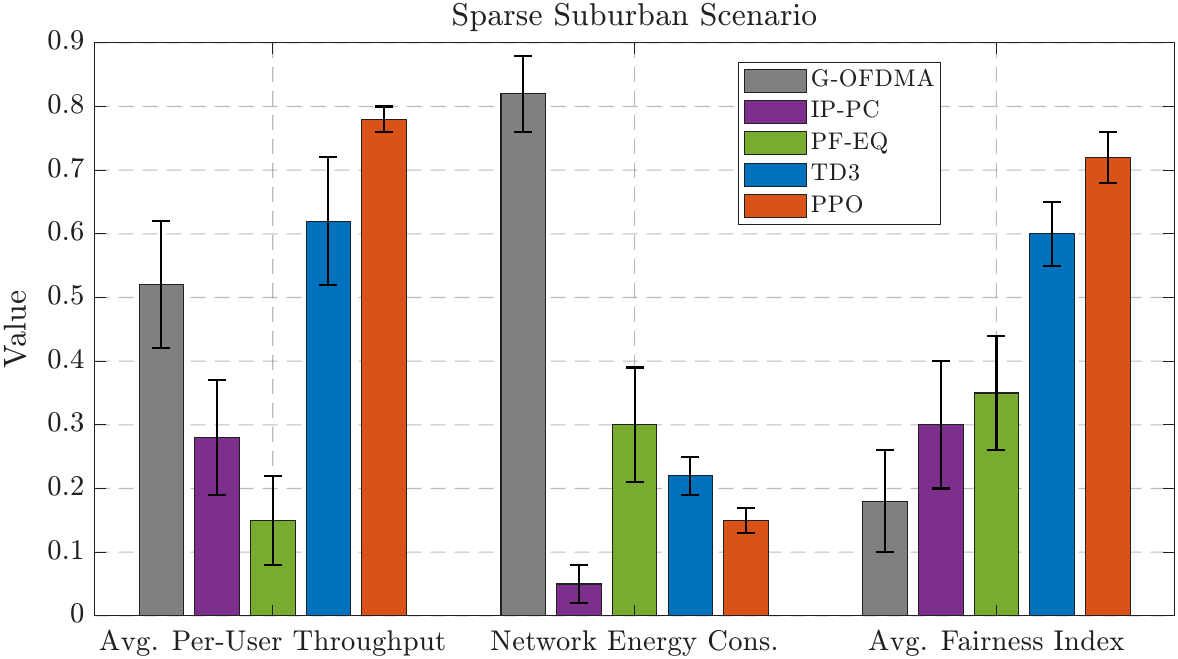}
        \caption{}
        \label{fig:sparse-suburban}
    \end{subfigure}
    \begin{subfigure}{0.32\textwidth}
        \centering
        \includegraphics[width=\textwidth]{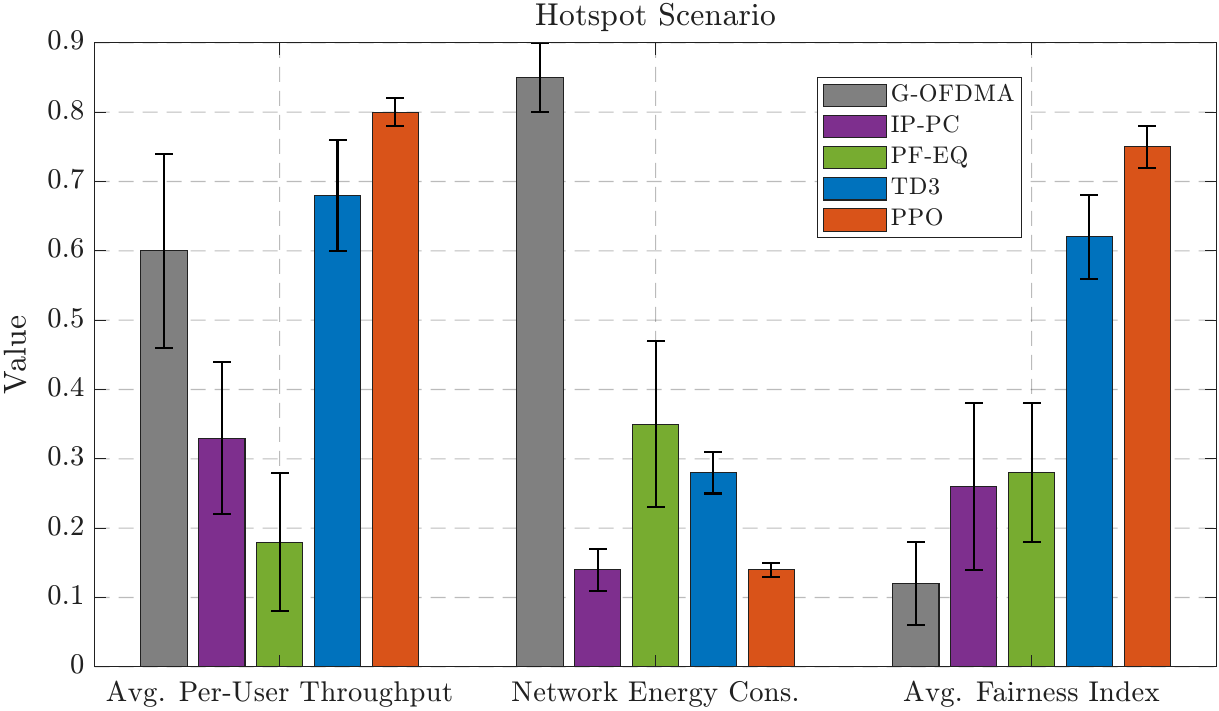}
        \caption{}
        \label{fig:hotspot-scenario}
    \end{subfigure}
    \vspace{0.16cm}
    \begin{subfigure}{0.32\textwidth}
        \centering
        \includegraphics[width=\textwidth]{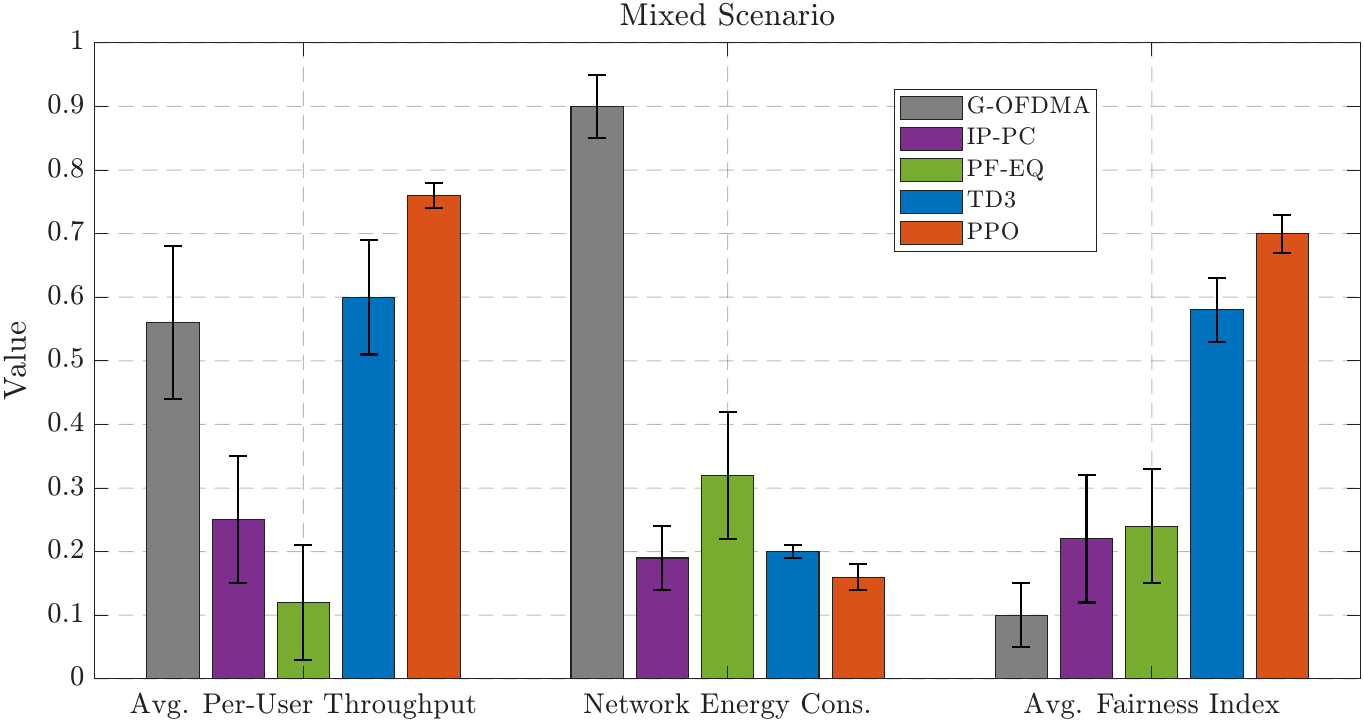}
        \caption{}
        \label{fig:mixed-scenario}
    \end{subfigure}
    \begin{subfigure}{0.32\textwidth}
        \centering
        \includegraphics[width=\textwidth]{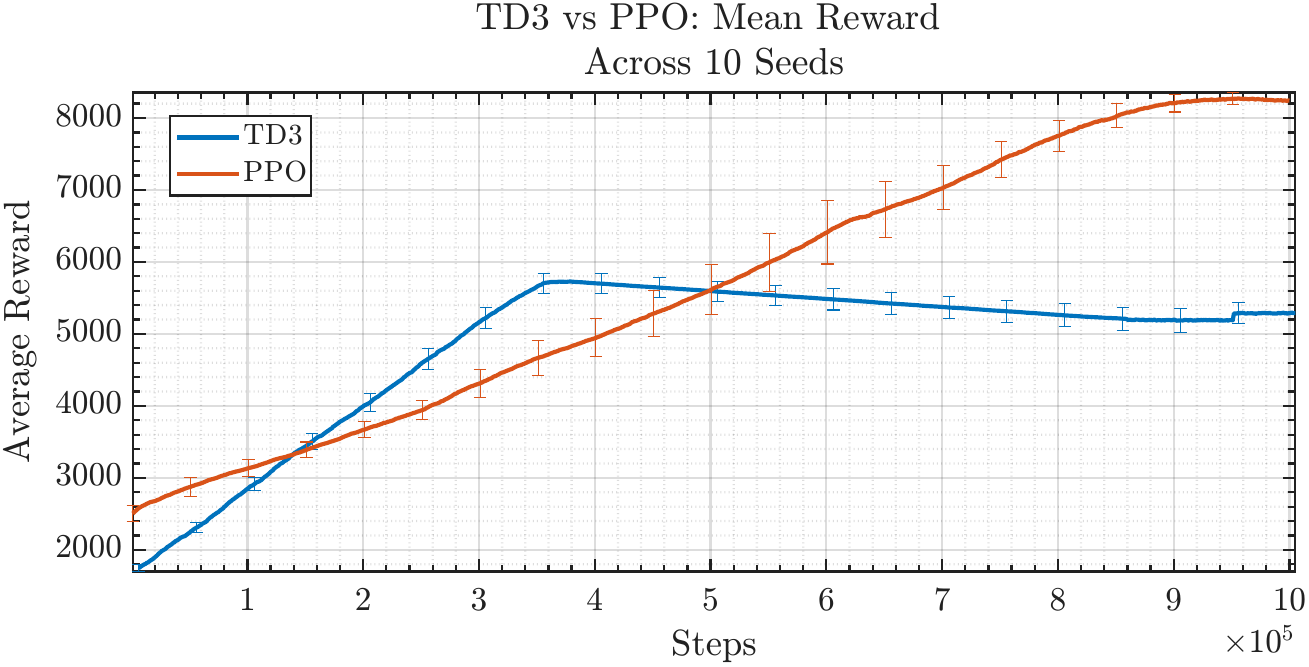}
        \caption{}
        \label{fig:reward-plot}
    \end{subfigure}
    \begin{subfigure}{0.32\textwidth}
        \centering
        \includegraphics[width=\textwidth]{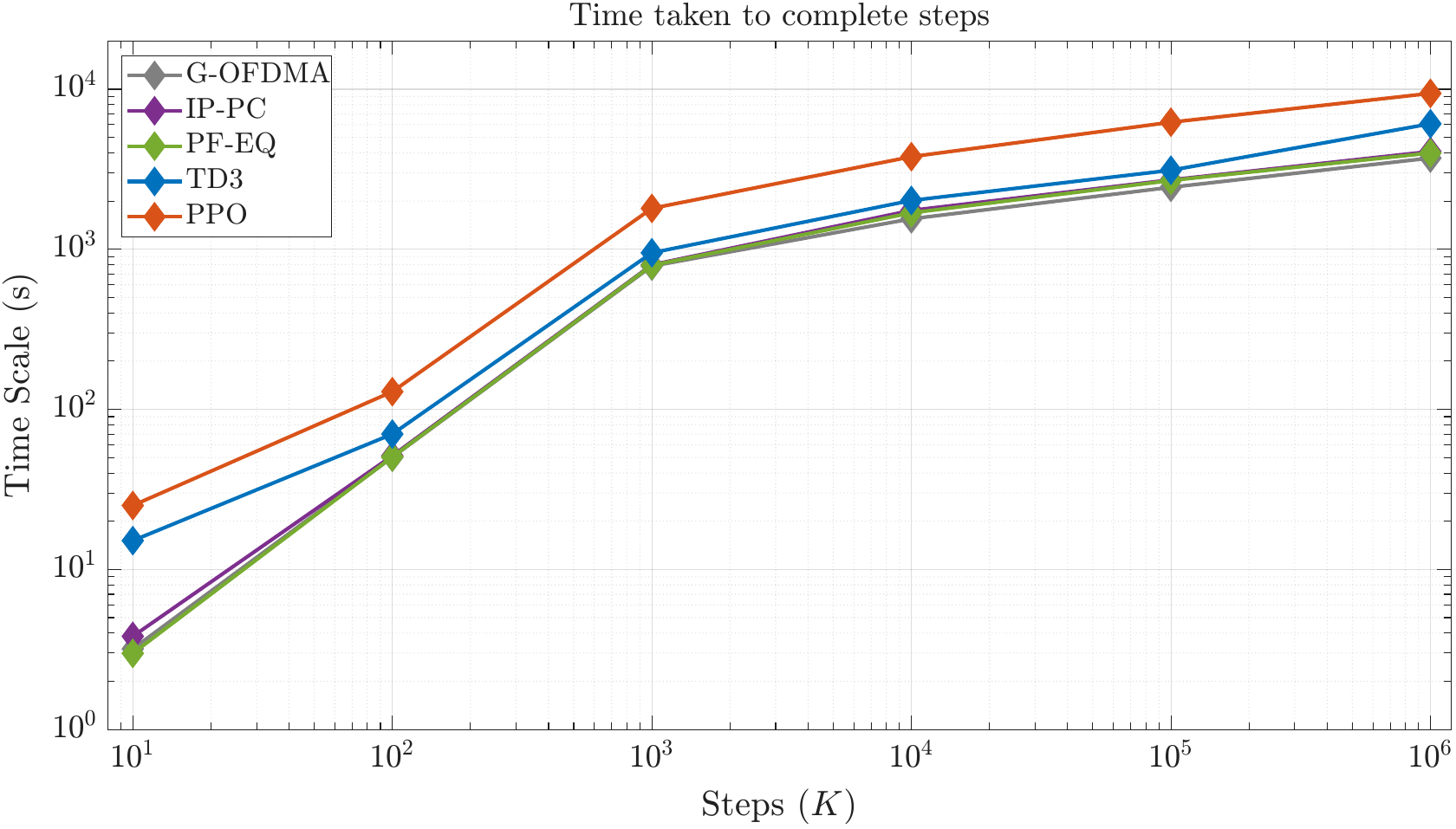}
        \caption{}
        \label{fig:time_scale}
    \end{subfigure}
    \caption{Comprehensive performance evaluation of the DRL-based O-RAN xApps against heuristic baselines. (a) Dense Urban Scenario: PPO significantly reduces network energy consumption by mitigating cross-tier interference. (b) Sparse Suburban Scenario: PPO achieves near-optimal fairness comparable to PF-EQ while maintaining high throughput. (c) Hotspot Scenario: DRL agents successfully balance load in high-traffic clusters. (d) Mixed Scenario: demonstrating policy robustness to randomised user distributions. (e) Mean reward convergence over 1M steps; PPO demonstrates superior stability compared to TD3. (f) Computational time complexity; the xApp inference latency remains within the Near-RT RIC tolerance window (\(10ms - 1s\)). Error bars represent the \(95\%\) confidence interval.}
    \label{fig:performance-metrics}
\end{figure*}

Fig.~\ref{fig:reward-plot} illustrates the training trajectory of the DRL agents. TD3 exhibits high sample efficiency, converging rapidly within the first \(3.5 \times 10^{5}\) steps. However, it suffers from instability and performance degradation in later stages, likely due to overestimation of values in the complex interference landscape. In contrast, PPO demonstrates a stable, monotonic ascent, ultimately achieving a significantly higher mean reward.

Fig.~\ref{fig:time_scale} quantifies the computational overhead. The heuristic baselines (G-OFDMA, IP-PC) operate in near-real-time (\(< 10^{-1}\)s per batch). The DRL inference times are orders of magnitude higher, with PPO being the most computationally intensive. However, the inference latency remains within the \(10ms - 1s\) window, validating the deployment of these agents as Near-RT RIC xApps rather than real-time MAC schedulers.\\

The results indicate that while TD3 offers faster initial deployment, PPO is the superior candidate for the RIC xApp. It provides a robust policy that maximises aggregate utility, successfully protecting cell-edge users (high fairness) and reducing the carbon footprint (low energy use) without compromising network capacity.

Regarding space complexity, the neural network architectures for PPO and TD3 comprise lightweight multi-layer perceptrons (MLPs) requiring minimal memory overhead (typically $<$ 10MB). This easily satisfies the stringent memory constraints of O-RAN Near-RT RIC controllers, which handle multiple concurrent messages.

\section{Conclusion}\label{conclusion}
In this paper, we addressed the resource orchestration problem in O-RAN HetNets by comparing PPO and TD3-based xApps. Our findings, based on realistic network topologies, reveal that while TD3 converges faster initially, PPO achieves a significantly higher overall reward by learning more effective policies for energy conservation and user fairness. This highlights a critical trade-off: TD3 is a sample-efficient algorithm suitable for rapid adaptation, whereas PPO’s methodical exploration yields a more globally optimal policy for performance-critical, energy-constrained environments. Future work will focus on extending this framework to distributed multi-agent DRL architectures (such as MAPPO or MADDPG) to address scalability, broadening the benchmark comparisons against a wider range of DRL baselines, and incorporating the effects of high-speed user mobility.

\section*{Acknowledgement}\label{acknow}
We thank Claude Formanek for his initial assistance with the algorithm design.

\bibliographystyle{IEEEtran}
\bibliography{references}

\end{document}